\documentclass[11pt]{article}

\usepackage[]{acl}

\usepackage{times}
\usepackage{latexsym}

\usepackage[T1]{fontenc}

\usepackage[utf8]{inputenc}

\usepackage{microtype}

\usepackage{inconsolata}

\usepackage{graphicx}

\usepackage{amsmath}
\usepackage{amssymb}
\usepackage{amsfonts}
\usepackage{mathrsfs}          

\usepackage{graphicx}          
\usepackage{booktabs}          
\usepackage{multirow}          
\usepackage{arydshln}          
\usepackage{makecell}          
\usepackage{array}             
\usepackage{adjustbox}         
\usepackage{subcaption}        
\usepackage[font=small]{caption} 
\usepackage{colortbl}          
\usepackage{pifont}            

\usepackage{enumitem}          

\usepackage{algorithm}
\usepackage{algpseudocode}

\usepackage{url}
\usepackage[hang,flushmargin]{footmisc}
\usepackage{cleveref}          

\usepackage{wasysym}           
\usepackage[most]{tcolorbox}   
\usepackage{nicefrac}          

\definecolor{grey}{RGB}{128,138,135}

\title{Towards Anticipatory Agents for Streaming Video Understanding}

\author{
  \textbf{Haolin Yang}\textsuperscript{1}\thanks{\ Equal contribution.} \quad
  \textbf{Feilong Tang}\textsuperscript{1}\footnotemark[1] \quad
  \textbf{Lingxiao Zhao}\textsuperscript{2}\footnotemark[1] \quad
  \textbf{Xinlin Zhuang}\textsuperscript{1} \quad
  \textbf{Yifan Lu}\textsuperscript{1} \\
  \textbf{Xiang An}\textsuperscript{3} \quad
  \textbf{Ming Hu}\textsuperscript{1} \quad
  \textbf{Xiaofeng Zhang}\textsuperscript{4} \quad
  \textbf{Abdalla Swikir}\textsuperscript{1} \quad
  \textbf{Junjun He}\textsuperscript{2} \\
  \textbf{Zongyuan Ge}\textsuperscript{5} \quad
  \textbf{Muhammad Haris Khan}\textsuperscript{1} \quad
  \textbf{Imran Razzak}\textsuperscript{1}\thanks{\ Corresponding author.} \\[4pt]
  \mdseries
  \textsuperscript{1}MBZUAI \quad
  \textsuperscript{2}Shanghai AI Laboratory \quad
  \textsuperscript{3}DeepGlint \\
  \textsuperscript{4}Shanghai Jiao Tong University \quad
  \textsuperscript{5}Monash University \\[2pt]
}

\begin{document}
\maketitle
\begin{abstract}
Real-time streaming video understanding requires continuous perception, proactive decision 
making, and responsive interaction with dynamically evolving visual content, posing challenges beyond conventional offline video processing. However, existing methods rely on alternating perception-reaction or asynchronous triggers, lacking \textit{task-driven planning} and \textit{future anticipation}, which limits their real-time responsiveness in evolving video streams. To this end, we propose a \textbf{StreamAgent} that anticipates the temporal intervals and spatial regions expected to contain future task-relevant information to enable proactive and goal-driven responses. Specifically, we integrate question semantics and historical observations through prompting the anticipatory agent to anticipate the temporal progression of key events, align current observations with the expected future evidence, and subsequently adjust the perception action (\textit{e.g.,} attending to task-relevant regions or continuously tracking in subsequent frames). To enable efficient inference, we design a \textit{streaming KV-cache memory} mechanism that constructs a hierarchical memory structure for selective recall of relevant tokens, enabling efficient semantic retrieval while reducing the overhead of storing all tokens in the traditional KV-cache. Extensive experiments on streaming and long video understanding tasks demonstrate that our method outperforms existing methods in response accuracy and real-time efficiency, highlighting its practical value for real-world streaming scenarios.
\end{abstract}

\section{Introduction}
\begin{figure*}[t]
  \centering
  \setlength{\abovecaptionskip}{0.1cm}
  \includegraphics[width=\textwidth]{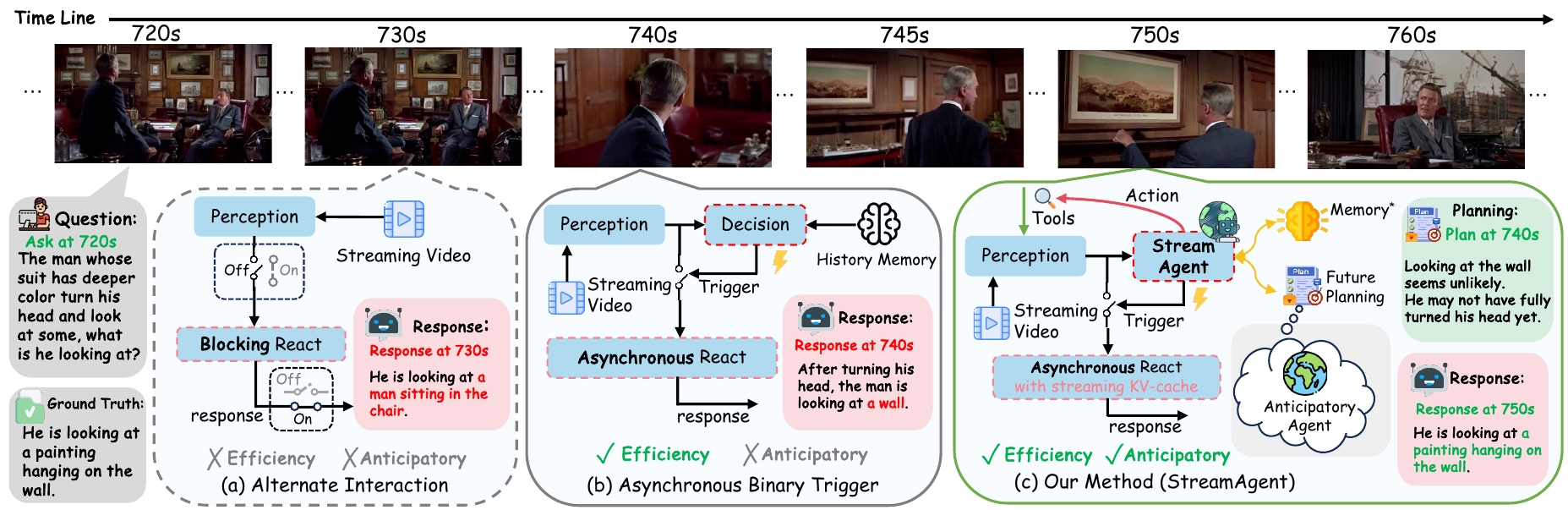}
  \caption{\textbf{Comparison between StreamAgent and existing methods.} Prior online methods~\cite{videollm-online, qian2025dispider} enable proactive interactions but rely on either (a) alternating interaction, causing slow processing, or (b) asynchronous binary triggers, leading to inaccurate responses. In contrast, \textbf{StreamAgent} integrates continuous perception with task-driven planning and future anticipation, enabling proactive identification of key temporal and spatial cues, and supporting efficient asynchronous reaction through the \textbf{streaming KV-cache}.}
  \label{intro}
\end{figure*}

As vision models are increasingly deployed in real-world scenarios (\textit{e.g.,} autonomous driving and intelligent surveillance), research has progressively shifted towards understanding continuous video streams. Streaming video understanding requires continuous processing of incoming frames, efficient extraction of critical information, proactive decision-making, and responsive interaction with dynamically evolving visual content, guided by explicitly defined queries.


Recent works such as VideoStreaming~\cite{qian2024streaming}, Flash-VStream~\cite{flashvstream}, and ReKV~\cite{di2025streaming} have explored multimodal large language models (MLLMs)~\cite{touvron2023llama,openai2023gpt4} for streaming video understanding, leveraging memory-based mechanisms to process long video streams and support real-time scene understanding. However, these models lack a proactive response mechanism to determine whether to respond immediately or continue observing. Therefore, We argue that beyond continuous perception, the key missing capability in existing streaming systems is \textit{anticipation}, the ability to proactively predicting and actively acquiring future task-relevant information, rather than passively consuming incoming frames.

To enable proactive interaction, VideoLLM-online~\cite{videollm-online} adopts an alternating ``perception--reaction'' mechanism (Fig.~\ref{intro}a), where a single LLM handles both perception and generation. However, the autoregressive nature of LLMs prevents parallel execution, delaying frame processing and reducing responsiveness. Dispider~\cite{qian2025dispider} addresses this by decoupling decision and reaction with an asynchronous binary trigger (Fig.~\ref{intro}b), yet still lacks task-driven planning and future anticipation. As a result, it prematurely responds with incomplete evidence (\textit{e.g.,} predicting ``\texttt{wall}'' instead of ``\texttt{painting}''), a failure rooted in insufficient temporal reasoning.

In this work, we propose a \textbf{StreamAgent} that anticipates the temporal intervals and spatial regions expected to contain future task-relevant information to enable proactive and goal-driven responses, as illustrated in Fig.~\ref{intro} (c). 
Specifically, the agent integrates question semantics with historical observations to predict the temporal progression and spatial locations of key events, aligning the current observation with the anticipated trajectory to determine whether sufficient information has been accumulated to trigger a response. If not, StreamAgent proactively refines its perception strategy (e.g., attending to task-relevant regions or persistently tracking targets across subsequent frames). As new video streams arrive, the agent iteratively updates its spatiotemporal focus to gather sufficient evidence for accurate responses, while an incremental memory update mechanism continuously integrates new information to ensure timely and coherent responses throughout the stream.

To address the long-context bottleneck in streaming inference~\cite{di2025streaming}, we further propose a \textbf{streaming KV-cache} that constructs a hierarchical memory structure for adaptive context retrieval. Each video clip is encoded into KV-cache via chunk-wise incremental prefill and offloaded to CPU as long-term memory. Upon a query, relevant entries are dynamically retrieved via layer-adaptive selection based on attention patterns, ensuring that only semantically pertinent content is recalled while alleviating GPU memory constraints.

We further validate our method through zero-shot evaluation on both streaming and offline video understanding benchmarks. In summary, our key contributions are as follows:
\begin{itemize}[leftmargin=*,itemsep=0pt]
\item We propose \textbf{StreamAgent} for streaming video understanding that anticipates future task-relevant temporal intervals and spatial regions to enable proactive responses.
\item We propose a \textbf{streaming KV-cache} that constructs a hierarchical memory structure, designed to enable efficient semantic retrieval and \textit{selective recall} of relevant tokens.
\item Experiments demonstrate that our method achieves state-of-the-art performance on streaming and long video understanding benchmarks.
\end{itemize}

\begin{figure*}[t]
  \centering
  \setlength{\abovecaptionskip}{0cm}
  \includegraphics[width=0.98\textwidth]{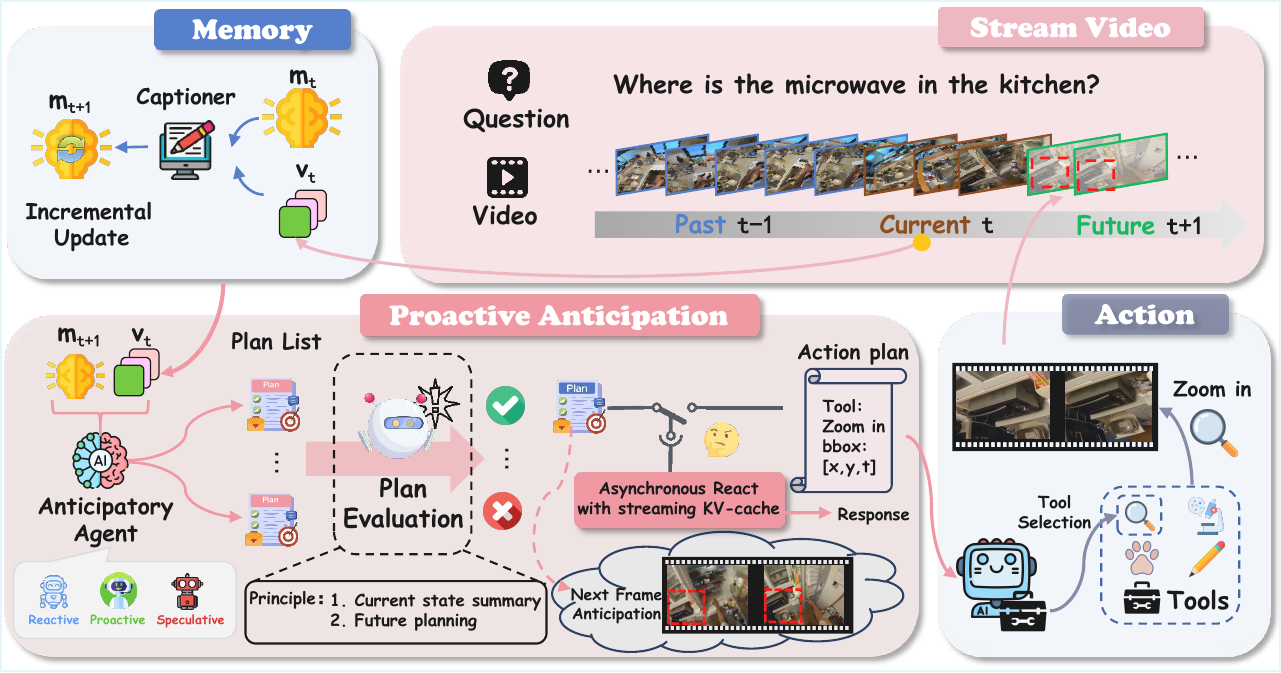}
  \caption{
  \textbf{Overview of the StreamAgent framework:} At each timestep, the system incrementally updates memory based on the current streaming video clip $v_t$ and memory $m_t$. Conditioned on the updated memory and current observation, the planner generates multiple future-aware plans through proactive anticipation simulating incoming videos, spanning three perspectives, \textit{Reactive, Proactive, and Speculative}. These plans are evaluated jointly considering current situational awareness and predicted future utility. The selected plan either triggers asynchronous reactive behavior or initiates proactive information hunting by invoking tools (\textit{e.g.,} zooming in task-relevant regions in subsequent frames), enabling goal-directed and proactive responses.
  }
  \label{figure:overview}
\end{figure*}

\section{Related Work}

\noindent\textbf{Streaming Video Understanding.}
Unlike traditional video understanding benchmarks~\cite{carreira2017quo, alamri2019audio, egoschema} that operate on sampled clips, streaming video understanding requires real-time perception, proactive decision-making, and responsive interaction over long video inputs.
Early efforts such as VideoLLM-Online~\cite{videollm-online} and Flash-VStream~\cite{flashvstream} explore online models that process frames via memory-based mechanisms.
To further disentangle the pipeline, Dispider~\cite{qian2025dispider} introduces an asynchronous binary trigger that separates decision from reaction.
More recent studies extend streaming video understanding along complementary axes, including streaming dialogue frameworks~\cite{chatterjee2025memoryefficientstreamingvideollmsrealtime, xiong2025streamingvideounderstandingmultiround, liu2025streamchatchattingstreamingvideo, chen2025livecclearningvideollm}, continuous perception under evolving content~\cite{wu2024videollmmodefficientvideolanguagestreaming, zhang2024internlmxcomposer25omnilivecomprehensivemultimodallongterm, wang2025streambridgeturningofflinevideo}, and temporal multi-turn interaction benchmarks~\cite{yang2025svbenchbenchmarktemporalmultiturn, wu2024streambenchbenchmarkingcontinuousimprovement, li2025videoscanenablingefficientstreaming}.
However, these methods either block perception during generation or rely on binary triggers without task-driven planning, limiting their ability to anticipate task-relevant information.
 
\noindent\textbf{Agentic Frameworks for Video Understanding.}
Recent studies have explored agentic approaches that leverage MLLMs to tackle complex video understanding tasks~\cite{zhi2025videoagent2, kugo2025videomultiagents, yuan2025videodeepresearch, chen2025lvagent, jeoung2024adaptive, zhang2024omagentmultimodalagentframework}.
These methods typically lack the capacity for forward-looking anticipation needed in real-time streaming scenarios.
In contrast, our StreamAgent integrates continuous perception with anticipatory planning and tool-augmented active perception, enabling goal-directed and efficient responses over streaming video without exhaustive frame-level processing.
 
\noindent\textbf{KV-Cache Optimization for Video Understanding.}
Processing long video streams incurs substantial 
KV-cache overhead. Recent methods address this through 
KV-cache 
compression~\cite{ning2025livevlm,chatterjee2025memoryefficientstreamingvideollmsrealtime}, 
visual feature 
compression~\cite{zhang2024flash,wu2024videollmmodefficientvideolanguagestreaming}, 
or attention-based token 
pruning~\cite{schneider2025quickvideo,di2025streaming}. 
However, these approaches rely on static attention 
patterns for token selection, limiting adaptability 
across transformer layers. Our streaming KV-cache
adopts a layer-adaptive retrieval strategy, allowing each layer to 
retrieve a variable number of KV entries tailored to its 
attention behavior.

\section{Methodology}

\subsection{Overview of StreamAgent Framework}
To tackle the challenges of Streaming VideoQA, we propose StreamAgent, a decision agent that anticipates future temporal intervals and spatial regions to contain task-relevant information, enabling proactive responses to user queries, as illustrated in Fig.~\ref{figure:overview}. Details of the proposed decision agent are elaborated in the following sections.

\begin{figure*}[t]
\centerline{\includegraphics[width=\linewidth]{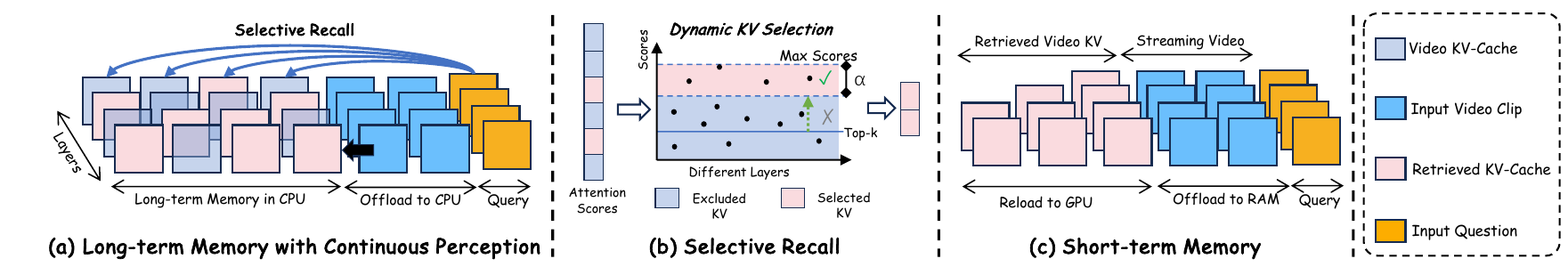}}
\setlength{\abovecaptionskip}{-0.1cm}
\caption{
\textbf{Overview of streaming KV-caches.} \textbf{(a) Long-term Memory:} As the video stream is continuously encoded, key-value (KV) caches from earlier frames are offloaded to CPU memory as long-term memory. \textbf{(b) Selective Recall:} Upon a query, relevant KV-caches are dynamically selected across layers based on attention scores. \textbf{(c) Short-term Memory:} Selected KV-caches are reloaded to GPU and combined with streaming inputs as short-term memory for efficient response generation.}

\label{fig:framework}
\end{figure*}


\noindent\textbf{Problem Formulation.} Streaming 
VideoQA~\cite{ding2025streammindunlockingframerate} 
requires answering proactive, context-sensitive, and 
real-time questions over video 
streams~\cite{ning2024infmllmefficientstreaminginference,guo2025seed15vltechnicalreport}. 
Let $\mathcal{V}^T = \{v_t\}_{t=1}^T$ denote a video 
stream up to time $T$, where $v_t$ is the $t$-th clip, 
and let $Q_t$ denote a question posed at timestamp $t$. 
The goal is to leverage the current clip $v_t$ and past 
video $\mathcal{V}_{1:t-1}$ to answer $Q_t$. When 
$\mathcal{V}_{1:t}$ is insufficient, the model should 
proactively defer and respond at a future timestamp 
$t'$ ($t \leq t' \leq T$) once $\mathcal{V}_{t+1:t'}$ 
provides sufficient evidence. We define a decision 
function $D(Q_t, \mathcal{V}_{1:t})$ that determines 
whether to wait or respond, and a response function $f$ 
that is invoked only when sufficient information has 
been accumulated.

\noindent\textbf{Incremental Memory Update.}
Continuously encoding all incoming video frames results 
in linearly growing visual tokens, which is prohibitive 
for real-time decision making. To address this, we compress the 
visual token sequence 
$\mathcal{X}_t = \{x_i\}_{i=1}^{N_t}$ of each incoming 
clip $v_t$ into a compact textual caption 
$\mathcal{C}_t = \{c_i\}_{i=1}^{n_t}$ ($N_t \gg n_t$), 
and maintain a memory state $m_t$ that 
is recurrently updated as:
\begin{equation}
\label{eq:memory}
m_t = \mathcal{U}(m_{t-1},\, \mathcal{C}_t),
\end{equation}
where $\mathcal{U}(\cdot)$ denotes the update function 
parameterized by the captioning model. This recurrence 
ensures that only the previous state $m_{t-1}$ and the 
current caption $\mathcal{C}_t$ are needed at each step, 
bounding memory cost at a constant level regardless of 
stream length. We note that this compression inevitably 
loses early fine-grained details; this limitation is 
compensated by the streaming KV-cache 
(Sec.~\ref{sec:method}), which preserves original visual 
tokens for selective recall during response generation.

\noindent\textbf{Proactive Anticipation with Plan Evaluation.}
Within the StreamAgent framework, we introduce the 
anticipatory stream agent, which leverages a lightweight 
MLLM, denoted as $A(\cdot)$, to 
forecast the temporal dynamics and spatial distribution 
of task-relevant events. The current state is represented 
as $S_t = A(m_t, v_t)$, integrating the memory $m_t$ 
(text) and the current observation $v_t$ (video).
From $S_t$, the agent reasons about potential future 
event trajectories $E_t^{(h)} = \{e_i\}_{i=t+1}^{t+h}$, 
where each $e_i$ denotes a predicted event at future 
timestamp $i$ and $h$ conceptually represents the 
temporal reasoning depth. To elicit diverse candidate 
plans that cover different levels of temporal foresight, 
StreamAgent prompts $A(\cdot)$ from three complementary 
perspectives. The \textit{Reactive} perspective 
($h{=}0$) grounds decisions solely in the current state 
$S_t$ without forecasting, favoring immediate response 
when evidence appears sufficient. The \textit{Proactive} 
perspective ($h{=}\delta$) extrapolates from current 
observations to anticipate near-future task-relevant 
events. The \textit{Speculative} perspective 
($h{=}\Delta,\;\Delta \gg \delta$) explores long-range 
future possibilities for queries requiring distant 
evidence.
Each perspective produces one or more candidate plans, 
forming:
\begin{equation}
\label{eq:plan}
P_j = A\!\left(E_t^{(h_j)},\, S_t \mid m_t,\, v_t
\right), \quad \mathcal{P} = \{P_j\}_{j=1}^{k},
\end{equation}
where $h_j \in \{0, \delta, \Delta\}$ and $k$ is the 
total number of candidate plans across all perspectives.

After generating the set of candidate plans, the same agent $A(\cdot)$ evaluates each plan using a separate scoring prompt, following the LLM-as-a-judge paradigm~\cite{zheng2023judging}. The evaluation is guided by the complementary criteria of assessing whether the currently accumulated evidence is sufficient to answer the query $Q$ and whether further observation of future video frames is likely to provide critical missing information.Based on these considerations, the agent produces a holistic assessment of each plan and selects the one deemed most appropriate for the current decision.

\noindent\textbf{Tool-Augmented Action.} Beyond passive frame consumption, StreamAgent acts as a goal-driven information explorer that proactively determines \textit{when}, \textit{where}, and \textit{how} to acquire critical visual evidence. Given the selected plan $\hat{P}$, the current state $S_t$, and the query $Q$, the anticipatory agent $A(\cdot)$ outputs a structured action specification indicating which tool to invoke and the corresponding spatial-temporal parameters (\textit{e.g.,} a bounding box for region-of-interest cropping or a target region for continuous tracking). Formally, at each timestamp $t$, the agent selects a subset of tools $\mathcal{T}'_t \subseteq \mathcal{T}$ and applies each $\phi_j \in \mathcal{T}'_t$ to a designated sub-region $v^j_{t+1} \subseteq v_{t+1}$ of the incoming frame, producing intermediate results $R_j = \phi_j(v^j_{t+1})$. The perception state is then updated as: \begin{equation} S_{t+1} = A\!\left(m_t,\, \{R_j\}_{j=1}^{|\mathcal{T}'_t|}\right). \end{equation} Crucially, tool invocations are not one-shot: the plan $\hat{P}$ may instruct the agent to \textit{persistently monitor} specific spatial regions or track designated objects across future frames $v_{t+1}, v_{t+2}, \dots$ until the target evidence is captured. By iteratively refining tool usage and perception targets along the predicted planning trajectory $\hat{P}$, StreamAgent exhibits proactive information hunting behavior, dynamically adapting its sensory focus to accumulate task-relevant evidence in streaming environments.

\subsection{Streaming Video KV-caches Memory Mechanism}
\label{sec:method}
This section presents the two processes, \textit{Continuous Perception} and \textit{Selective Recall}, that constitute the streaming KV-caches memory mechanism, as illustrated in Fig.~\ref{fig:framework}.

\noindent\textbf{Continuous Perception.}
Continuous Perception adopts an incremental encoding and prefill strategy, storing the resulting KV caches from each video clip into long-term memory for future retrieval (\textit{i.e.,} Selective Recall).
As shown in Fig.~\ref{fig:framework}(a), 
each video clip is sequentially encoded and prefilled, producing a KV-cache:
$\{H_\text{clip}^k,H_\text{clip}^v\}= \{(\mathbf{k}_j, \mathbf{v}_j)\}_{j=1}^{l_P},$ 
where $l_P$ denotes the total KV length from prior clips.
To minimize peak activation memory usage, we apply chunked prefill within each clip.
Given an incoming video clip $v_i$ with token sequence 
$\mathbf{Z}^{v_i} = \{ z_j^{v_i} \}_{j=1}^n,$ we divide it into $C$ chunks 
$
\mathbf{C}^{v_i} = \{\mathbf{Z}^{v_i}_j\}^{\mu}_{j=1},
$
where $\mu$ and $n$ denote the total number of chunks and tokens in $v_i$, respectively, and each token $z_j^{v_i} \in \mathbb{R}^h$ has hidden size $h$.
At each transformer layer, we sequentially prefill the current visual tokens chunk 
$X=\{z_{j+l_C}\}_{j=1}^{l_X}$
while maintaining the corresponding  KV-cache of the past chunks as $\{H_\text{chunk}^k,H_\text{chunk}^v\}= \{(\mathbf{k}_j, \mathbf{v}_j)\}_{j=1}^{l_C}$, where $l_X$ represents the chunk size, and $l_C$ denotes the length of the previously accumulated KV cache in the current clip.
The attention keys and values are then built as:
\begin{equation}
\small
\label{eq:QKV_linear_transformation}
K = [H_\text{clip}^k, H_\text{chunk}^k, X  W_K], 
V = [H_\text{clip}^v, H_\text{chunk}^v, X  W_V],
\end{equation}
\noindent where $W_K$, and $W_V$ are the $K$ and $V$ weight matrices, stored for future retrieval, respectively. 


%
\begin{table*}[t]
    \small
    \centering
    \setlength{\abovecaptionskip}{0.2cm}
    \begin{adjustbox}{max width=2.1\columnwidth}
    \begin{tabular}{l@{}c@{\hspace{3pt}}|ccccccc|cccc|cccc|c}
    \toprule
    \multirow{2}{*}{\textbf{Model}} & \multirow{2}{*}{\textbf{\#Frames}} & 
    \multicolumn{7}{c|}{\textbf{Real-Time Visual Perception}} & 
    \multicolumn{4}{c|}{\textbf{Backward Tracing}} & 
    \multicolumn{4}{c|}{\textbf{Forward Active Responding}} & \multirow{2}{*}{\textbf{Overall}} \\
    \addlinespace[2pt]
    \cmidrule[0.5pt](lr){3-9} \cmidrule[0.5pt](lr){10-13} \cmidrule[0.5pt](lr){14-17}
    \addlinespace[2pt]
    &  & OCR & ACR & ATR & STU & FPD & OJR & Avg. & 
    EPM & ASI & HLD & Avg. & 
    REC & SSR & CRR & Avg. & Avg. \\
    \midrule
    \midrule
    Human Agents & - & 94.0 & 92.6 & 94.8 & 92.7 & 91.1 & 94.0 & 93.2 & 92.6 & 93.0 & 91.4 & 92.3 & 95.5 & 89.7 & 93.6 & 92.9 & 92.8 \\
    \midrule
    \multicolumn{18}{c}{\textbf{Proprietary Multimodal Models}} \\
    \midrule
    Gemini 1.5 Pro~\cite{gemini} & 1fps & 87.3 & 67.0 & 80.2 & 54.5 & 68.3 & 67.4 & 70.8 & 68.6 & 75.7 & 52.7 & 62.3 & 35.5 & 74.2 & 61.7 & 57.2 & 65.3 \\
    GPT-4o~\cite{gpt4o} & 64 & 69.1 & 65.1 & 65.5 & 50.0 & 68.3 & 63.7 & 63.6 & 49.8 & 71.0 & 55.4 & 58.7 & 27.6 & 73.2 & 59.4 & 53.4 & 58.6 \\
    \midrule
    \multicolumn{18}{c}{\textbf{Open-source Offline VideoLLMs}} \\
    \midrule
    \textcolor{grey}{LLaVA-NeXT-Video-7B~\cite{llava_next_video}} & \textcolor{grey}{64} & \textcolor{grey}{69.8} & \textcolor{grey}{59.6} & \textcolor{grey}{66.4} & \textcolor{grey}{50.6} & \textcolor{grey}{72.3} & \textcolor{grey}{61.4} & \textcolor{grey}{63.3} & \textcolor{grey}{51.2} & \textcolor{grey}{64.2} & \textcolor{grey}{9.7} & \textcolor{grey}{41.7} & \textcolor{grey}{34.1} & \textcolor{grey}{67.6} & \textcolor{grey}{60.8} & \textcolor{grey}{54.2} & \textcolor{grey}{53.1} \\
    \textcolor{grey}{LLaVA-OneVision-7B\cite{llava_onevision}} & \textcolor{grey}{64} & \textcolor{grey}{67.1} & \textcolor{grey}{58.7} & \textcolor{grey}{69.8} & \textcolor{grey}{49.4} & \textcolor{grey}{71.3} & \textcolor{grey}{60.3} & \textcolor{grey}{62.8} & \textcolor{grey}{52.5} & \textcolor{grey}{58.8} & \textcolor{grey}{23.7} & \textcolor{grey}{45.0} & \textcolor{grey}{24.8} & \textcolor{grey}{66.9} & \textcolor{grey}{60.8} & \textcolor{grey}{50.9} & \textcolor{grey}{52.9} \\
    \textcolor{grey}{Qwen2-VL-7B~\cite{Qwen2VL}} & \textcolor{grey}{64} & \textcolor{grey}{69.1} & \textcolor{grey}{53.2} & \textcolor{grey}{63.8} & \textcolor{grey}{50.6} & \textcolor{grey}{66.3} & \textcolor{grey}{60.9} & \textcolor{grey}{60.7} & \textcolor{grey}{44.4} & \textcolor{grey}{66.9} & \textcolor{grey}{34.4} & \textcolor{grey}{48.6} & \textcolor{grey}{30.1} & \textcolor{grey}{65.7} & \textcolor{grey}{50.8} & \textcolor{grey}{48.9} & \textcolor{grey}{52.7} \\
    \textcolor{grey}{InternVL-V2-8B~\cite{chen2024internvl}} & \textcolor{grey}{64} & \textcolor{grey}{68.5} & \textcolor{grey}{58.7} & \textcolor{grey}{69.0} & \textcolor{grey}{44.9} & \textcolor{grey}{67.3} & \textcolor{grey}{56.0} & \textcolor{grey}{60.7} & \textcolor{grey}{43.1} & \textcolor{grey}{61.5} & \textcolor{grey}{27.4} & \textcolor{grey}{44.0} & \textcolor{grey}{25.8} & \textcolor{grey}{57.6} & \textcolor{grey}{52.9} & \textcolor{grey}{45.4} & \textcolor{grey}{50.1} \\
    \textcolor{grey}{LongVU-7B~\cite{shen2024longvu}} & \textcolor{grey}{1fps} & \textcolor{grey}{55.7} & \textcolor{grey}{49.5} & \textcolor{grey}{59.5} & \textcolor{grey}{48.3} & \textcolor{grey}{68.3} & \textcolor{grey}{63.0} & \textcolor{grey}{57.4} & \textcolor{grey}{43.1} & \textcolor{grey}{66.2} & \textcolor{grey}{9.1} & \textcolor{grey}{39.5} & \textcolor{grey}{16.6} & \textcolor{grey}{69.0} & \textcolor{grey}{60.0} & \textcolor{grey}{48.5} & \textcolor{grey}{48.5} \\

    \midrule
    \multicolumn{18}{c}{\textbf{Open-source Online VideoLLMs}} \\
    \midrule
    Flash-VStream-7B~\cite{flashvstream} & 1fps  & 25.5 & 32.1 & 29.3 & 33.7 & 29.7 & 28.8 & 29.9 & 36.4 & 33.8 & 5.9 & 25.4 & 5.4 & \textbf{67.3} & \textbf{60.0} & \underline{44.2} & 33.2 \\
    VideoLLM-online-8B~\cite{videollm-online} & 2fps & 8.1 & 23.9 & 12.1 & 14.0 & 45.5 & 21.2 & 20.8 & 22.2 & 18.8 & \underline{12.2} & 17.7 & - & - & - & - & - \\
    Dispider~\cite{qian2025dispider} & 1fps  & 57.7 &\underline{49.5} &62.1 & \underline{44.9} & 61.4 & 51.6 & 54.5 & 48.5 & 55.4 & 4.3 & 36.1 & 18.0 & 37.4 & 48.8 & 34.7 & 41.8 \\
     TimeChat-Online-7B~\cite{timechatonline} & 1fps  &\underline{69.8}  &48.6  &\textbf{64.7}  &\underline{44.9}  & \textbf{68.3}  & \underline{55.4}  & \underline{58.6}  &\underline{53.9}  &\textbf{62.8}  &9.1  &\textbf{42.0}  &\underline{32.5}  &36.5  &40.0  &36.4  & \underline{45.6} \\
    \midrule
        \textbf{StreamAgent-7B (Ours)} & 1fps & \textbf{71.2} & \textbf{53.2} & \underline{63.6} & \textbf{53.9} & \underline{67.3} & \textbf{58.7} & \textbf{61.3} & \textbf{54.8} & \underline{58.1} & \textbf{25.8} & \underline{41.7} & \textbf{35.9} & \underline{48.4} & \underline{52.0} & \textbf{45.4} & \textbf{49.4}  \\
    \bottomrule

    \end{tabular}
    \end{adjustbox}
    \caption{Evaluation results on OVO-Bench~\cite{li2025ovobenchfarvideollmsrealworld} comprising three categories: i) \textit{Real-Time Visual Perception} (OCR: Optical Character Recognition, ACR: Action Recognition, ATR: Attribute Recognition, STU: Spatial Understanding, FPD: Future Prediction, OJR: Object Recognition), ii) \textit{Backward Tracing} (EPM: Episodic Memory, ASI: Action Sequence Identification, HLD: Hallucination Detection), and iii) \textit{Forward Active Responding} (REC: Repetition Event Count, SSR: Sequential Steps Recognition, CRR: Clues Reveal Responding).}
    \label{tab:ovo-bench}
\end{table*}

\noindent\textbf{Selective Recall. }
%
%
Selective Recall dynamically retrieves question-relevant 
KV-caches from long-term memory into GPU-resident 
short-term memory, adapting to the attention behavior of 
each transformer layer for efficient and accurate 
inference.

\begin{itemize} [leftmargin=0pt, labelsep=-5pt]
\item \textit{\quad Retrieving with Dynamic Attention Pattern. }
Recent advances in long-sequence modeling with LLMs show that the number of influential keys and values varies across different layers~\cite{lee2024infinigen}.
Therefore, unlike ReKV~\cite{di2025streaming}, which retains a fixed number of KV entries per layer, we propose a layer-adaptive retrieval strategy that dynamically selects variable numbers of KV entries based on attention patterns (see Fig.~\ref{fig:framework}(b)). This allows layers with broader attention to access more entries, while layers with narrower focus retrieve fewer.
In detail, during video stream prefill, we first compute a representative feature per frame to enable rapid scoring: $\frac{1}{T_f} \sum_{j=1}^{T_f} \mathbf{k}_j \in \mathbb{R}^{D'}$, where $T_f$ is the number of tokens per frame, and $\mathbf{k}_j$ is the $j$-th key vector. 
Subsequently, to match frame relevance to the question, we compute the query-level attention descriptor by averaging over all question tokens: $
\frac{1}{T_q} \sum_{j=1}^{T_q} \mathbf{q}_j \in \mathbb{R}^{D'}
$, 
where $T_q$ is the number of query tokens, and $\mathbf{q}_j$ is the score vector of the $j$-th token across each attention head.
Since softmax exponentially amplifies higher attention scores, we filter out less relevant frames by retaining only those with scores within a margin $\alpha$ of the maximum, \textit{i.e.,} $\text{score} \geq \max - \alpha$.
Thus, the set of important keys can be formulated as:
\begin{equation}
\label{dynamic}
\mathcal{J}_h = \left\{ j \in \{1, \dots, L_k\} \mid \max_{j'}(\mathbf{S}_h) - \mathbf{S}_{h,j} \leq \alpha \right\},
\end{equation}
where $\mathbf{S}_h \in \mathbb{R}^{L_k}$ and $L_k$ denote the attention scores for head $h$ and total number of stored keys, respectively.
For example, consider $l$ frames, with $n$ frames having attention scores above the threshold $\alpha$. The softmax-normalized importance of the $i$-th frame is given by:
\begin{equation}
\beta_i = 
\frac{e^{\mathbf{S}_{h,i}}}{\sum_{j=1}^l e^{\mathbf{S}_{h,j}}},
\end{equation}
where $\mathbf{S}_{h,i}$ denote the attention scores of $i$-th frame.
If $\alpha = 6$ and the score of the $i$-th frame falls below $\max(\mathbf{S}_h) - 6$, its softmax weight $\beta_i$ is at most $1/e^6 \approx 1/403.4$ of the maximum. Such negligible weights justify its exclusion, contributing minimally to the final attention distribution.
\begin{figure*}[!t]
  \centering
  \begin{tabular}{cc}
    \includegraphics[width=\linewidth]{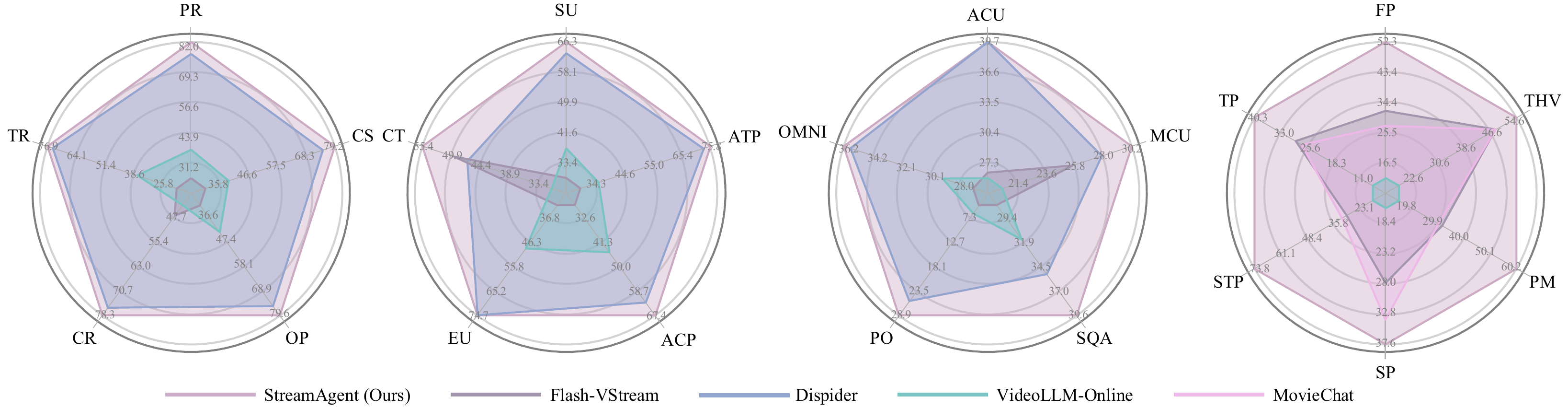} 
  \end{tabular}
  \setlength{\abovecaptionskip}{0.2cm}
  \caption{Comparative radar plots of StreamAgent and existing online video LLMs on diverse benchmarks, StreamingBench (first three) and OV-Bench (last), demonstrating StreamAgent's superior unified video representation capabilities. 
  }
  \vspace{-0.2cm}
  \label{fig:stream}
\end{figure*}
\item \textit{\quad Answering with Retrieved KV.}
The retrieved KV-caches serve as the contextual input for video understanding. As illustrated in Fig. \ref{fig:framework}(c), the KV entries corresponding to the current question are concatenated with the KV-caches retrieved from the long-term memory. In this case, the $K$ and $V$ used for attention computation are represented as:
\begin{equation}
\label{retrieval_run}
K = [HX^k_\text{out}, X_q \cdot W_K], \quad
V = [H^v_\text{out}, X_q \cdot W_V],
\end{equation}
where $H_\text{out}^k$ and $H_\text{out}^v$ represent the key and value cache outputs which are retrieved from the long-term memory, respectively, 
and $X_q$ denotes either the current input question or the next token that will be generated.
\end{itemize}

\section{Experiments}
\label{sec:experiments}

\subsection{Implementation Details} 
We employ the lightweight Qwen2.5VL-3B~\cite{bai2025qwen25vltechnicalreport} as the agent's core for planning and tool coordination, and the larger Qwen2.5VL-7B~\cite{bai2025qwen25vltechnicalreport} for precise interactions. Input video frames are resized to 224×224 at 1 FPS, following Dispider~\cite{qian2025dispider}. For long-term memory, we first store the KV-cache in GPU memory, and offload to CPU memory once a predefined threshold is reached. The default value of $\alpha$ is set to 3. All experiments are conducted on NVIDIA A800 (80GB) GPUs with FP16 precision. Further setup details are provided in the Appendix C.

\begin{figure*}
    \centering
    \includegraphics[width=\linewidth]{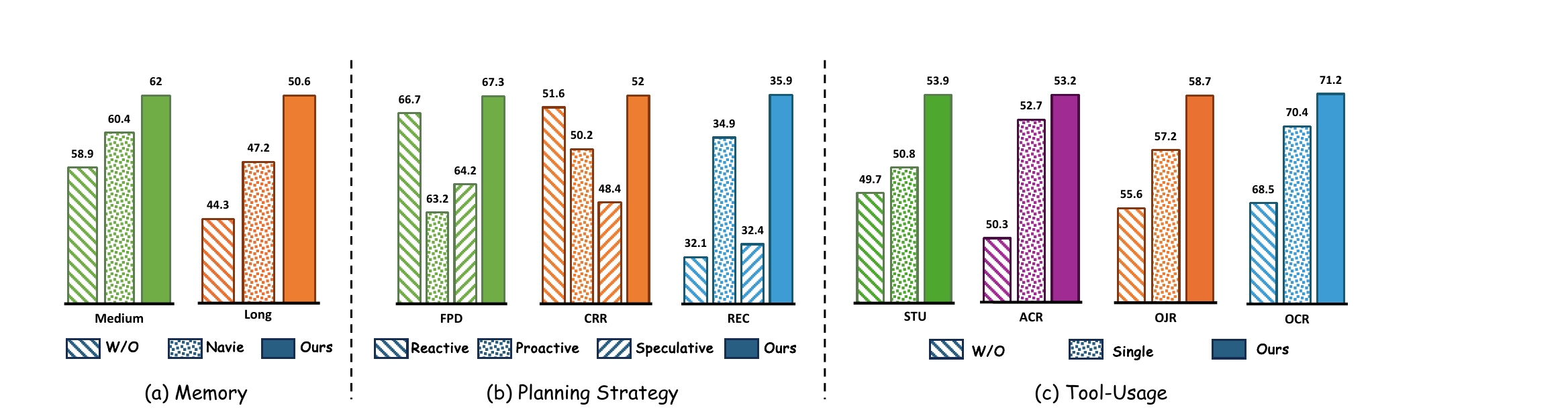}
    \setlength{\abovecaptionskip}{-0.2cm}
    \caption{\textbf{Ablation Study} on three core components: (a) The effectiveness of the memory module is evaluated on the \textit{medium} and \textit{long} subsets of the VideoMME benchmark. (b) The impact of different planning strategies is evaluated on an OVOBench~\cite{li2025ovobenchfarvideollmsrealworld} subset designed to test temporal foresight. (c) Tool usage is assessed on a separate subset to examine how proactive interaction enhances performance.}
    \label{fig:agent}
    \vspace{-0.2cm}
\end{figure*}

\subsection{Results on Streaming Video Benchmarks}
\label{sec:exp_stream}
We evaluate StreamAgent on three streaming video benchmarks tailored to different aspects of streaming video understanding, including StreamingBench~\cite{lin2024streamingbench}, OVO-Bench~\cite{li2025ovobenchfarvideollmsrealworld} and OVbench~\cite{huang2025onlinevideounderstandingovbench}, under strict real-time settings. 
Table~\ref{tab:ovo-bench} reports StreamAgent’s performance on OVO-Bench which consists of 12 diverse tasks grouped into three core capabilities: Backward Tracing, Real-Time Visual Perception, and Forward Active Responding. StreamAgent achieves the highest overall performance among online models and even outperforms most offline. Notably, it improves Forward Active Responding by 10.7\% over Dispider~\cite{qian2025dispider}, and closely matches the top offline model, highlighting the effectiveness of proactive planning and streaming-aware design.
To further assess generalization, we evaluate on StreamingBench and OVBench. While StreamingBench focuses on real-time perception with fine-grained, time-stamped queries, OVBench emphasizes long-range temporal reasoning across 16 subtasks spanning past, present, and future. As illustrated in Fig.~\ref{fig:stream}, StreamAgent achieves state-of-the-art results on both benchmarks, significantly outperforming previous online models and approaching offline baselines.
These gains are driven by our proactive StreamAgent, which accurately predicts task-relevant temporal and spatial cues, and our efficient streaming KV-cache mechanism, which enhances long-form understanding by selective recall while reducing computational overhead.

\subsection{Ablation Study}
\label{sec:exp_ablations}

\noindent\textbf{Effectiveness of Agent Design.}
To evaluate the contribution of each core component in our StreamAgent framework, we conduct a comprehensive ablation study along three key axes: \textit{incremental memory, planning strategy, and tool-based active perception}. Each experiment is designed to isolate the effect of the corresponding module on the final performance.

\noindent\textit{1. Memory.} Specifically, we compare the performance of our memory mechanism against approaches without memory and those using traditional segmentation-based captioning, where captions are generated for each segment and then concatenated. Our method shows significant gains on longer videos (Fig.~\ref{fig:agent} (a)), which confirms that Markov memory not only reduces redundancy but also enhances long-term context modeling crucial for high-quality video question answering.

\noindent\textit{2. Planning.} For planning, we compare \textit{Reactive, Proactive, and Speculative}.
strategies with our heuristic-based planning. As shown in Fig.~\ref{fig:agent} (b), our approach consistently outperforms fixed strategies across all time scales. Notably, the performance of speculative planning remains relatively limited across videos of all lengths, likely due to the inherent uncertainty of long-horizon predictions. Additionally, the nature of the dataset may contribute to this trend, as many questions tend to favor short-term reasoning and do not require extensive temporal foresight.

\begin{figure}[!t]
  \centering
  \begin{tabular}{cc}
    \includegraphics[width=\linewidth]{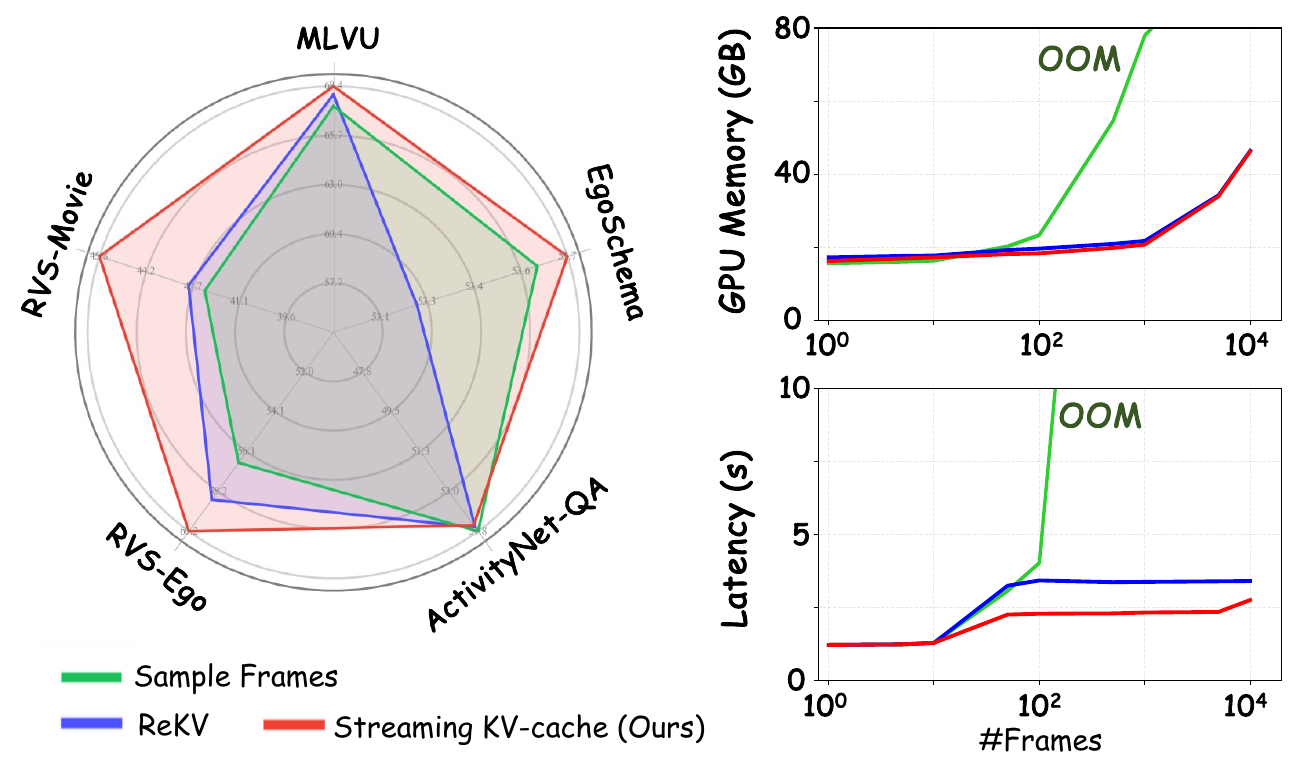} 
  \end{tabular}
  \caption{Streaming KV-cache boosts Streaming VideoQA accuracy and achieves over 30\% faster inference than ReKV when evaluated with Qwen2.5vl-7B on an A800 (80GB).}
  \vspace{-0.2cm}
  \label{fig:kv}
\end{figure}

\noindent\textit{3. Tool-Use.} Finally, we ablate external tools to evaluate the impact of proactive information hunting. As shown in Fig.~\ref{fig:agent} (c), tool-based perception significantly improves accuracy with minimal overhead, underscoring the value of targeted, strategic perception beyond passive observation.

\begin{table}[t]
    \small
    \centering
    \begin{adjustbox}{max width=1\columnwidth}
    \begin{tabular}{lc|ccccc}
        \toprule
        \textbf{Model} & \textbf{\#Frames} & \textbf{MLVU} & \multicolumn{4}{c}{\textbf{VideoMME}} \\
        \cmidrule(lr){4-7}
        & & & Short & Medium & Long & Overall \\
        \midrule
        VideoAgent & 87 & 57.8 & 63.6 & 55.4 & 49.0 & 56.0 \\ VideoMemAgent & 72 & 58.2 & 55.3 & \textbf{64.2} & \textbf{52.7} & 57.4 \\
        ReAgent-V & 35 & \underline{60.7} & \textbf{73.5} & 58.2 & 49.8 & \underline{60.7} \\
        \midrule
        \textbf{StreamAgent-7B} & 1fps & \textbf{67.2} & \underline{73.4} & \underline{62.0} & \underline{50.6} & \textbf{62.9} \\
        \bottomrule
    \end{tabular}
    \end{adjustbox}
    \caption{Ablation of different video understanding agents.}
    \label{tab:video_agent}
    \vspace{-0.4cm}
\end{table}

\begin{table}[t]
    \small
    \centering
    \setlength{\abovecaptionskip}{0cm}

    \begin{adjustbox}{max width=1\columnwidth}
    \begin{tabular}{lc|cccc}
    \toprule
    \multirow{2}{*}{\textbf{Model}} & \multirow{2}{*}{\textbf{\#Frames}} & \multirow{2}{*}{\textbf{MLVU}} & \multirow{2}{*}{\textbf{LongVideoBench}} & \multicolumn{2}{c}{\textbf{VideoMME}} \\
    \cmidrule(lr){5-6}
    \addlinespace[1pt]
    & & & & overall & long \\
    \midrule
    \textbf{Video Length} & - & 3$\sim$120 min & 8 sec$\sim$60 min & 1$\sim$60 min & 30$\sim$60 min \\
    \midrule
    \multicolumn{6}{c}{\textbf{Open-Source Offline VideoLLMs}} \\
    \midrule
    LLaMA-VID-7B & 1fps & 33.2 & - & - & - \\
    MovieChat-7B & 2048 & 25.8 & - & 38.2 & 33.4 \\
    LLaVA-Next-Video-7B & 32 & - & 43.5 & 46.6 & - \\
    VideoChat2-7B & 16 & 47.9 & 39.3 & 39.5 & 33.2 \\
    LongVA-7B & 128 & 56.3 & - & 52.6 & 46.2 \\
    Qwen2.5-VL-7B & 1fps  & 66.9 & 61.5 & 63.2 & 50.4 \\
    \midrule
    \multicolumn{6}{c}{\textbf{Open-source Online  VideoLLMs}} \\
    \midrule
    Dispider-7B & 1fps & 61.7 & - & 57.2 & - \\
    VideoChat-Online-8B & 2fps & - & - & 52.8 & 44.9 \\
    TimeChat-Online-7B & 1fps  & 65.4 & 57.7 & 62.5 & 49.2 \\
    \midrule
    \textbf{StreamAgent-7B} & 1fps & \textbf{67.2} & \textbf{57.9} & \textbf{62.9} & \textbf{50.6} \\
    \bottomrule

    \end{tabular}
\end{adjustbox}

\caption{Results on offline long video benchmarks. We report the accuracy on the MLVU~\cite{mlvu}, LongVideobench~\cite{wu2024longvideobenchbenchmarklongcontextinterleaved} and VideoMME~\cite{videomme} without subtitles. }
 \label{tab:long_video_benchmarks}
\end{table}

\begin{figure*}[t]
\setlength{\abovecaptionskip}{0cm}
\centerline{\includegraphics[width=\linewidth]{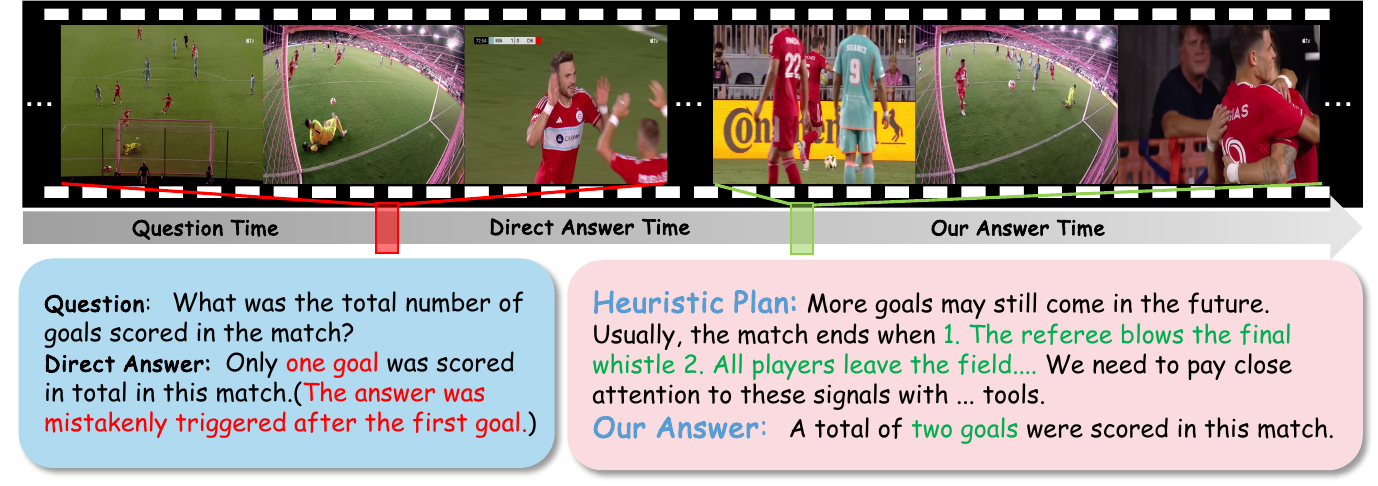}}
\caption{
\textbf{Proactive Anticipation examples.}
The video stream is processed frame by frame.
On the left, a Direct Answer is triggered prematurely right after the first goal, resulting in an incorrect response due to the lack of future planning.
On the right, the proposed Heuristic Plan waits for clear signs that the game has ended, such as the referee's final whistle, before answering. 
}
\label{fig:visualization}
\vspace{-0.2cm}
\end{figure*}

\noindent\textbf{Effectiveness of Streaming KV-cache.}
To evaluate the effectiveness of our streaming KV-cache mechanism, we follow the experimental protocol of ReKV~\cite{di2025streaming} and test on five diverse benchmarks covering different kinds of traditional scenarios: \textit{MLVU}~\cite{mlvu}, \textit{EgoSchema}~\cite{egoschema}, \textit{ActivityNet-QA}~\cite{activitynet_qa}, \textit{RVS-Ego} and \textit{RVS-Movie}~\cite{flashvstream}.
We compare against representative baselines: Sampled Frames, and ReKV. As shown in Fig.~\ref{fig:kv}, our hybrid architecture enables efficient communication, achieving up to 30\% lower latency than ReKV as frame counts increase. Moreover, GPU memory usage remains on par with Sampled Frames, while our method consistently outperforms baselines. More details can be found in Appendix C.

\noindent\textbf{Ablation Study on Video Understanding Agent.}
In the offline setting, we further compare the performance of various video understanding agents, VideoAgent~\cite{wang2024videoagent}, VideoMemAgent~\cite{fan2024videoagentmemoryaugmentedmultimodalagent} and ReAgent-V~\cite{zhou2025reagent}. Benefiting from our memory mechanism and streaming KV-cache, our method processes videos at 1 FPS and shows outperformance (Table~\ref{tab:video_agent}) on video benchmarks.

\noindent\textbf{Results on Offline Long Video Tasks.}
In addition to the streaming setting, we also evaluate StreamAgent on three offline long-form video
understanding benchmarks: VideoMME~\cite{videomme}, MLVU~\cite{mlvu}, and
LongVideoBench~\cite{wu2024longvideobenchbenchmarklongcontextinterleaved}. In the offline setting, the entire video is provided as input to the VideoLLMs. Table~\ref{tab:long_video_benchmarks} demonstrates that StreamAgent exhibits superior offline video understanding capabilities compared to recent state-of-the-art VideoLLMs, including LLaMA-VID~\cite{llamavid}, MovieChat~\cite{song2024moviechat}, LLava-next-video~\cite{llava_next_video}, VideoChat~\cite{li2023videochat}, LongVA~\cite{longva}, Qwen2.5-VL~\cite{Qwen2VL}, Dispider~\cite{qian2025dispider}, VideoChat-Online~\cite{huang2025onlinevideounderstandingovbench}. Thanks to incremental memory module and KV cache retrieval mechanism, StreamAgent handles conventional scenario well and achieves competitive performance with offline models.


\noindent\textbf{Case Study.}
To evaluate StreamAgent's capability in \textit{Proactive Event Anticipation}, we present a case from StreamingBench~\cite{lin2024streamingbench} dataset involving
the question “What was the total number of
goals scored in the match?”, a scene from a football match, as shown in Fig.~\ref{fig:visualization}. When the model lacked a plan, it prematurely triggered a response after the first goal and incorrectly answered “1 goal.”. Through heuristic planning, we avoid such premature answers by using tools to focus on key signals and responding after those signals are observed. This case highlights the value of future-aware planning and tool integration in proactive video understanding.

\section{Conclusion}
In this paper, we propose StreamAgent for streaming video understanding, which proactively anticipates the temporal intervals and spatial regions of future task-relevant information, enabling proactive and accurate responses. Furthermore, we introduce streaming KV-cache that constructs a hierarchical memory structure and supports selective recall of relevant tokens for efficient semantic retrieval. Experiments demonstrate that our proposed method achieves state-of-the-art performance on streaming and long video understanding benchmarks in both response accuracy and real-time efficiency.

\section*{Limitations}
Although our proposed framework and dataset significantly enhance the streaming capabilities of existing offline video large language models (Video-LLMs), there are still some noteworthy limitations.

First, although the decision model is compact and decoupled from the main language model, its decision quality is still constrained by the capacity of the small activation model. While tool usage helps mitigate this issue, the model may still struggle to accurately predict the optimal response timing in highly complex or ambiguous scenarios.

Second, while StreamAgent improves response accuracy through collaborative interactions among multiple agents and proactive acquisition of future information, the lack of large-scale streaming data for training remains a challenge that limits its full potential.

Third, although our streaming KV-cache effectively handles long-term memory, we aim to further reduce computational overhead and improve efficiency by removing redundant information before it enters the MLLM, through various redundancy elimination techniques.

\bibliography{custom}

\appendix



\section*{Appendix Table of Contents}

\begin{itemize}[leftmargin=*,itemsep=0pt]

\item \textbf{A \quad Additional Related Work}

\item \textbf{B \quad Benchmark and Baseline}

\item \textbf{C \quad Memory Analysis of Streaming KV cache}

\item \textbf{D \quad Detail of Evaluation}

\item \textbf{E \quad Pseudocode}

\item \textbf{F \quad Future Work}

\end{itemize}


\section{Additional Related Work.}


\noindent\textbf{Long Video Understanding} 
Inspired by the powerful reasoning capabilities of LLMs~\cite{yu2025flowreasoningtrainingllms}, recent works have explored using LLMs to address complex video-related tasks. Since LLMs primarily process text, various methods~\citep{korbar2024textconditionedresamplerlongform,weng2024longvlmefficientlongvideo, zhang2023videollamainstructiontunedaudiovisuallanguage,jin2023chatunivi, wang2024efficienttemporalextrapolationmultimodal} have been developed to efficiently train multimodal projectors to connect the visual encoder and LLMs or leverage caption-centric information. 
With the success of VideoLLM~\cite{kwaikeyeteam2025kwaikeyevltechnicalreport,Shu_2025_CVPR,yan2025learningstreamingvideorepresentation,coreteam2025mimovltechnicalreport}, attention has shifted to the more complex task of long video understanding~\cite{wang2025videotreeadaptivetreebasedvideo,zhou2025glimpselargevisionlanguagemodels, wang2024videollamblongcontextvideounderstanding}, which is typically facilitated through token compression~\cite{pollard2025videotokensmerge,lee2024videotokenmerginglongform,choi2024vidtldrtrainingfreetoken} and frame selection~\cite{liu2025boltboostlargevisionlanguage,zhang2025qframequeryawareframeselection,hu2025mllmbasedvideoframe,huang2025fragframeselectionaugmented}.
This approach effectively reduces the token sequence length, allowing for better handling of longer videos. Recent work has enhanced the model's capabilities by using inference~\cite{yang2025scalingnoisescalinginferencetimesearch,feng2025videor1reinforcingvideoreasoning,li2024videovistaversatilebenchmarkvideo,fei2024videoofthoughtstepbystepvideoreasoning,tian2025egor1chainoftoolthoughtultralongegocentric,wu2025mmsearchr1incentivizinglmmssearch}, often searching through long videos to obtain more granular details and achieve more accurate grounding during video understanding.

\section{Benchmark and Baseline.}

\noindent\textbf{MLVU}$_\texttt{dev-mc}$~\citep{mlvu} is the multiple-choice subset of the MLVU-dev benchmark. It focuses on evaluating the long-form video understanding of MLMs. The question-answer pairs are manually labeled and can be divided into 3 groups: single-detail, multi-detail, and holistic. The evaluation metric is Accuracy.

\noindent\textbf{LongVideoBench}~\citep{wu2024longvideobenchbenchmarklongcontextinterleaved} is a video question-answering benchmark specifically designed for LMMs. It aims to evaluate LMMs' ability to process long, interleaved video and subtitle inputs up to an hour in length.
The benchmark includes: 3,763 varying-length web-collected videos with their subtitles, covering diverse themes. A novel "referring reasoning" video question-answering task, where questions contain referring queries that point to specific video contexts, requiring the model to reason over relevant video details. 6,678 human-annotated multiple-choice questions, categorized into 17 fine-grained categories.

\noindent\textbf{VideoMME}~\citep{videomme} is the first comprehensive, full-spectrum evaluation benchmark for LMMS specifically designed for video analysis. It addresses the current gap in assessing LMMs' ability to process sequential visual data, moving beyond their traditional focus on static image understanding. Here are its key features:
Diverse Video Types: Video-MME covers 6 primary visual domains and 30 subfields, ensuring broad applicability across various scenarios.
Varied Temporal Durations: It includes short, medium, and long videos, ranging from 11 seconds to 1 hour, to evaluate robust contextual understanding.
Broad Data Modalities: Beyond video frames, the benchmark integrates other modalities like subtitles and audio to assess the all-around capabilities of MLMs.
High-Quality Annotations: It uses rigorous manual labeling by expert annotators to ensure precise and reliable model assessment.

\noindent\textbf{EgoSchema}~\citep{egoschema} is a diagnostic benchmark for long VideoQA, featuring over 5000 multiple-choice questions and long temporal certificate length. It challenges AI models with long-term understanding, as current state-of-the-art models achieve significantly lower accuracy compared to human performance.

\noindent\textbf{ActivityNet-QA}~\citep{activitynet_qa} 
encompasses human-annotated QA pairs on 5,800 videos derived from the ActivityNet~\citep{activitynet} dataset. 
This benchmark is designed to assess the capabilities of VideoQA models in long-term spatiotemporal reasoning. Our evaluation methodology aligns with that of Video-ChatGPT~\citep{video_chatgpt}, employing \texttt{GPT-3.5-turbo-0613} to judge the accuracy of the open-ended VideoQA responses.

\begin{table*}[t]
\centering
\renewcommand{\arraystretch}{1.4}
\Huge
\fontseries{b}
\setlength{\arrayrulewidth}{1pt}
\begin{adjustbox}{max width=\textwidth}
\begin{tabular}{lcc|ccccccccccc|ccccc|ccccc|ccc|c}
\toprule
\multirow{2}{*}{\textbf{Model}} & \multirow{2}{*}{\textbf{Params}} & \multirow{2}{*}{\textbf{Frames}} & \multicolumn{11}{c|}{\textbf{Real-Time Visual Understanding}} & \multicolumn{5}{c|}{\textbf{Omni-Source Understanding}} & \multicolumn{5}{c|}{\textbf{Contextual Understanding}} & \multirow{2}{*}{\textbf{Overall}} \\
\cmidrule(lr){4-14} \cmidrule(lr){15-19} \cmidrule(lr){20-24}
 &  &  & OP & CR & CS & ATP & EU & TR & PR & SU & ACP & CT & \textbf{All}  & ER & SCU & SD & MA & \textbf{All} & ACU & MCU & SQA & PO & \textbf{All} \\\hline
\multicolumn{25}{c}{\textbf{Human}} \\\hline
Human{$^\ddag$} & - & - & 89.47 & 92.00 & 93.60 & 91.47 & 95.65 & 92.52 & 88.00 & 88.75 & 89.74 & 91.30 & 91.46  & 88.00 & 88.24 & 93.60 & 90.27 & 90.26  & 88.80 & 90.40 & 95.00 & 100 & 93.55 & 91.66\\\hline
\multicolumn{25}{c}{\textbf{Proprietary MLLMs}} \\\hline
Gemini 1.5 pro & - & 1 fps & 79.02 & 80.47 & 83.54 & 79.67 & 80.00 & 84.74 & 77.78 & 64.23 & 71.95 & 48.70 & 75.69 & 46.80 & 39.60 & 74.90 & 80.00 & 60.22 & 51.41  & 40.73 & 54.80 & 45.10 & 48.73 & 67.07 \\
GPT-4o & - & 64 & 77.11 & 80.47 & 83.91 & 76.47 & 70.19 & 83.80 & 66.67 & 62.19 & 69.12 & 49.22 & 73.28 & 41.20 & 37.20 & 43.60 & 56.00 & 44.50  & 41.20 & 38.40 & 32.80 & 56.86 & 38.70 &60.15\\
Claude 3.5 Sonnet & - & 20 & 80.49 & 77.34 & 82.02 & 81.73 & 72.33 & 75.39 & 61.11 & 61.79 & 69.32 & 43.09 & 72.44 & 31.60 & 34.00 & 32.80 & 48.80 & 36.80  & 38.40 & 34.80 & 34.40 & 64.71 & 37.70 & 57.68\\\hline
\multicolumn{25}{c}{\textbf{Open-Source Video MLLMs}} \\\hline
LLaVA-OneVision & 7B & 32 & 80.38 & 74.22 & 76.03 & 80.72 & 72.67 & 71.65 & 67.59 & 65.45 & 65.72 & 45.08 & 71.12 & 40.80 & 37.20 & 33.60 & 44.80 & 38.40  & 35.60 & 36.00 & 27.27 & 29.55 & 32.74 & 56.36\\
Qwen2-VL & 7B & 0.2-1 fps & 75.20 & 82.81 & 73.19 & 77.45 & 68.32 & 71.03 & 72.22 & 61.19 & 61.47 & 46.11 & 69.04 & 41.20 & 22.00 & 32.80 & 43.60 & 34.90  & 31.20 & 26.00 & 39.60 & 22.73 &  31.66& 54.14\\
MiniCPM-V 2.6 & 8B & 32 & 71.93 & 71.09 & 77.92 & 75.82 & 64.60 & 65.73 & 70.37 & 56.10 & 62.32 & 53.37 & 67.44& 40.80 & 24.00 & 34.00 & 41.20 & 35.00  & 34.00 & 31.60 & 41.92 & 22.22 & 34.97 & 53.85\\
LLaVA-NeXT-Video & 32B & 64 & 78.20 & 70.31 & 73.82 & 76.80 & 63.35 & 69.78 & 57.41 & 56.10 & 64.31 & 38.86 & 66.96 & 37.69 & 24.80 & 34.40 & 42.80 & 34.90  & 29.20 & 30.40 & 35.35& 18.18 & 30.79 & 52.77\\
InternVL-V2 & 8B & 16 & 68.12 & 60.94 & 69.40 & 77.12 & 67.70 & 62.93 & 59.26 & 53.25 & 54.96 & 56.48 & 63.72& 37.60 & 26.40 & 37.20 & 42.00 & 35.80 & 32.00 & 31.20 & 32.32 & 40.91 & 32.42 & 51.40\\
Kangaroo & 7B & 64 & 71.12 & 84.38 & 70.66 & 73.20 & 67.08 & 61.68 & 56.48 & 55.69 & 62.04 & 38.86 & 64.60 & 37.60 & 31.20 & 28.80 & 39.20 & 34.20 & 32.80 & 26.40 & 33.84 &  16.00& 30.06 & 51.10\\
LongVA & 7B & 128 & 70.03 & 63.28 & 61.20 & 70.92 & 62.73 & 59.50 & 61.11 & 53.66 & 54.67 & 34.72 & 59.96 & 39.60 & 32.40 & 28.00 & 41.60 & 35.40  & 32.80 & 29.60 & 30.30 & 15.91 & 29.95& 48.66\\
VILA-1.5 & 8B & 14 & 53.68 & 49.22 & 70.98 & 56.86 & 53.42 & 53.89 & 54.63 & 48.78 & 50.14 & 17.62 & 52.32& 41.60 & 26.40 & 28.40 & 36.00 & 33.10 & 26.80 & 34.00 & 23.23 & 17.65 & 27.35&43.20\\
Video-CCAM & 14B & 96 & 56.40 & 57.81 & 65.30 & 62.75 & 64.60 & 51.40 & 42.59 & 47.97 & 49.58 & 31.61 & 53.96 & 33.60 & 22.00 & 28.40 & 34.80 & 29.70 & 27.60 & 24.40 & 16.67 &  22.73& 22.88 &42.53 \\
Video-LLaMA2 & 7B & 32 & 55.86 & 55.47 & 57.41 & 58.17 & 52.80 & 43.61 & 39.81 & 42.68 & 45.61 & 35.23 & 49.52 & 30.40 & 32.40 & 30.40 & 36.00 & 32.40  & 24.80 & 26.80 & 18.67 & 0.00 & 21.93 & 40.40\\\hline
\multicolumn{25}{c}{\textbf{Streaming MLLMs}} \\\hline
Flash-VStream & 7B& -& 25.89& 43.57 &24.91 &23.87& 27.33& 13.08 &18.52& 25.20& 23.87 &48.70 &23.23  &25.91& 24.90& 25.60& 28.40& 26.00& 24.80& 25.20 &26.80 &1.96& 24.12 & 24.04\\
VideoLLM-online & 8B & 2 fps &  39.07 &40.06 &34.49& 31.05& 45.96 &32.40 &31.48& 34.16 &42.49 &27.89& 35.99 &31.20 &\textbf{26.51} &24.10 &32.00 &28.45 &24.19 &29.20 &30.80 &3.92 &26.55 & 32.48\\
Dispider  & 7B & 1 fps & 74.92& 75.53& 74.10& 73.08& 74.44& 59.92& 76.14& 62.91& 62.16& 45.80&67.63 &35.46 &25.26& 38.57 &43.34 &35.66& 39.62& 27.65& 34.80& 25.34 &33.61 & 53.12\\
\textbf{StreamAgent}  & 7B & 1 fps & \textbf{79.63} & \textbf{78.31}& \textbf{79.28} & \textbf{75.87} & \textbf{74.74} & \textbf{76.92} & \textbf{82.94} & \textbf{66.31} & \textbf{73.69} & \textbf{55.40} &\textbf{74.28} &\textbf{35.86} &26.26& \textbf{38.87} & \textbf{44.04} & \textbf{36.26}& \textbf{39.72}& \textbf{30.25}& \textbf{39.60}& \textbf{28.90} &\textbf{34.62} & \textbf{57.02}\\\bottomrule

\end{tabular}
\end{adjustbox}
\caption{Performance comparison on StreamingBench on Omni-source Understanding, Contextual Understanding, and Real-Time Visual Understanding. Omni-source Understanding includes Emotion Recognition (ER), Scene Understanding (SCU), Source Discrimination (SD), and Multimodal Alignment (MA). Contextual Understanding includes Misleading Context Understanding (MCU), Anomaly Context Understanding (ACU), Sequential Question Answering (SQA) and Proactive Output (PO). Real-Time Visual Understanding includes Object Perception (OP), Causal Reasoning (CR), Clips Summarization (CS), Attribute Perception (ATP), Event Understanding (EU), Text-Rich Understanding (TR), Prospective Reasoning (PR), Spatial Understanding (SU), Action Perception (ACP), and Counting (CT). Results are categorized into Human, Proprietary MLLMs, and Open-Source MLLMs for a comprehensive evaluation.}
\label{tab:streamingbench}
\end{table*}

\noindent\textbf{RVS-Ego} and \textbf{RVS-Movie}~\citep{flashvstream} are Streaming VideoQA benchmarks, constructed using 10 long videos from the Ego4D dataset~\citep{ego4d} and 22 long videos from the MovieNet dataset~\citep{movienet}, respectively. 
These benchmarks feature open-ended questions paired with timestamps, 
which are initially generated by GPT-4V~\citep{gpt4v} and GPT-4~\citep{gpt4}, and subsequently refined through manual filtering.

\noindent\textbf{OVO-Bench}~\citep{li2025ovobenchfarvideollmsrealworld} is a dataset specifically designed to evaluate Video Large Language Models (Video-LLMs) on their ability to understand online, streaming video content. Unlike traditional offline benchmarks that assess models after they have seen the entire video, OVO-Bench focuses on real-time reasoning, requiring models to answer questions at any given timestamp while the video is playing. OVO-Bench defines a comprehensive suite of tasks to evaluate the temporal reasoning and visual understanding capabilities of Video Large Language Models (Video-LLMs) in online settings. These tasks are grouped into three core categories. The first, Backward Tracing, assesses a model’s ability to recall and reason about past events, including retrieving key moments (EPM), identifying the correct sequence of actions (ASI), and detecting hallucinated responses to irrelevant questions (HLD). The second category, Real-Time Visual Perception, focuses on understanding the present visual context through spatial reasoning (STU), object recognition (OJR), attribute identification (ATR), action recognition (ACR), text recognition (OCR), and even predicting what might happen next (FPD). The third category, Forward Active Responding, goes beyond passive perception by requiring models to delay responses until sufficient evidence is available—such as recognizing repeated events (REC), detecting procedural transitions (SSR), and waiting for critical clues before answering (CRR). Together, these tasks provide a rigorous evaluation of how well a model can perceive, remember, and anticipate in dynamic, real-world video environments.

\noindent\textbf{StreamingBench}~\citep{lin2024streamingbench} introduces the first comprehensive benchmark tailored for evaluating how well multimodal large language models (MLLMs) understand streaming video in realistic scenarios. Comprising 900 videos with 4,500 human-curated QA pairs spread across 18 task types, it simulates continuous video interactions by posing five time-staggered questions per video. The benchmark assesses three key understanding dimensions: real-time visual understanding (recognizing objects, actions, or text in the moment), omni-source understanding (integrating synchronized visual and audio inputs), and contextual understanding (maintaining continuity across interactions, detecting anomalies, filtering misleading cues, and responding proactively based on preceding context).

\noindent\textbf{OVBench}~\citep{huang2025onlinevideounderstandingovbench} are Streaming VideoQA benchmarks, denoting the unique challenges and characteristics of online video understanding compared to traditional offline settings. First, it emphasizes the online temporal perspective, where questions are grounded in specific moments, past, present, or future, enabling much finer time sensitivity than offline approaches. It highlights how time-dependent contexts cause answers to evolve dynamically, meaning that the same question may yield different responses as the stream progresses. Third, it points to the need for real-time spatio-temporal interaction in applications like AR glasses or autonomous driving, where immediate and accurate responses to live environments are crucial.

To rigorously assess the performance of StreamAgent, we compare it against a broad array of baseline
models that span different capabilities and design philosophies. Proprietary closed-source models
include GPT-4o, known for its advanced multimodal reasoning, and Gemini-1.5-Pro,
optimized for long-context multimodal input. Among open-source video-language models, the
evaluation includes:

\noindent\textbf{VideoChat2}: A chat-centric video-language model built to support multi-turn, conversational interactions grounded in video content. It emphasizes interactive understanding, with capabilities for dialogue continuity, temporal grounding, and multi-modal alignment, making it ideal for assistive agents and education scenarios.

\noindent\textbf{MiniCPM-V}: MiniCPM-V is a lightweight, vision-language model developed for efficient multimodal understanding and generation tasks. Designed with a compact architecture, it achieves strong performance on image captioning, visual question answering, and other vision-language benchmarks, while maintaining low computational requirements. This makes MiniCPM-V ideal for deployment in resource-constrained environments such as mobile devices and edge computing scenarios.

\noindent\textbf{VILA-1.5}: VILA-1.5 is a state-of-the-art vision-language model that excels in multimodal understanding and generation tasks. It integrates visual and textual information through a unified architecture, enabling high performance on benchmarks such as image captioning, visual question answering, and image-text retrieval. With improved alignment between visual and linguistic modalities, VILA-1.5 offers both accuracy and efficiency, making it suitable for a wide range of real-world applications.

\noindent\textbf{Qwen2} and \textbf{Qwen2.5-VL} families: Developed by Alibaba, these models exhibit high performance on both image and video understanding benchmarks through advanced multi-modal alignment. Qwen2.5-VL particularly excels in handling long-context and dense visual-textual reasoning tasks, supported by a unified visual-language architecture.

\noindent\textbf{InternVL-2}: A high-performing open-source model that scales both the architecture and training data to improve generalization. It incorporates vision-language alignment techniques, dense region grounding, and optimized pretraining routines to support both short and long video tasks with high efficiency.

\noindent\textbf{Kangaroo}: Designed specifically for long-context video input, Kangaroo adopts hierarchical attention and memory-efficient tokenization strategies to process hundreds of frames. It is particularly effective for document-level video understanding, such as meeting summarization or sports analysis.

\noindent\textbf{Long-LLaVA}: A variant of LLaVA optimized for efficient multi-frame processing. It integrates temporal coherence constraints and cross-frame attention mechanisms, making it capable of capturing nuanced motion patterns and temporal dependencies for improved video QA and description.

\noindent\textbf{LLaVA‑OneVision}: LLaVA‑OneVision is an open-source large multimodal model that excels across single-image, multi-image, and video understanding tasks. Built on a SigLIP vision encoder and Qwen‑2 language backbone, it uses a unified visual token strategy—including AnyRes‑9 for high-res images and frame-level pooling for video—to balance representation across modalities.

\noindent\textbf{MovieChat}: MovieChat is an innovative long‑video understanding framework that combines vision foundation models and large language models using a dual-memory mechanism inspired by the Atkinson–Shiffrin model. It processes video frames with a sliding‑window tokenization approach and maintains a rapidly updated short‑term memory buffer along with a compact long‑term memory.

\noindent\textbf{LLaVA‑NeXT‑Video}: LLaVA‑NeXT‑Video is an open‑source video‑language assistant built by fine‑tuning the LLaVA‑NeXT image model (based on Qwen‑1.5 32B) with high‑quality video instruction data (\~830K samples) mixed with image data. It leverages the AnyRes representation—originally designed for high‑resolution images—to naturally process videos as sequences of frames, enabling strong zero‑shot video understanding ([huggingface.co][2]). With techniques like length generalization and DPO fine‑tuning, the model achieves top open‑source performance on benchmarks like Video‑MME and NextQA‑MC, while remaining efficient in inference and deployment.

\noindent\textbf{ReAgent-V}: ReAgent-V is an agent-based video-language framework that enhances large vision-language models with dynamic frame selection, multi-step reasoning, and self-correction. By integrating tool-using agents and a feedback-driven refinement process, it achieves superior performance on video understanding tasks with improved efficiency and accuracy.

\noindent\textbf{VideoAgent}:An agent framework for long-form video understanding that mimics human cognition through LLM-guided reasoning, CLIP-based retrieval, and VLM-driven state updates.

\noindent\textbf{VideoMemAgent}: An agent framework for structured long-form video understanding that integrates unified temporal and object memory with tool-augmented reasoning, enabling multi-round chain-of-thought inference across complex video content.

\section{Memory Analysis of Streaming KV cache}
In this section, we analyze the memory consumption of our approach, focusing on activation memory and key-value (KV) cache memory.

\noindent\textbf{Activation Memory Analysis. }
The activation memory of modern LLM architecture mainly comes from two components of each transformer layer: 1) Attention layer and 2) MLP layer. We analyze the potential activation memory usage in formulas in the followings and show that chunked-prefill can approximately reduce the activation memory by $C$ times, where $C$ is the number of chunks.

\noindent\textbf{\emph{Attention Layer.} }
For an inference computation, the input tensor of the $i$-th attention layer ($X^i$) is first layer-normalized to produce $X^i_a$, which is then fed into the attention module. 
The output of the attention module after a residual add and layer normalization forms the $X^i_O$. 
The $X^i_O$ can be calculated by:

\begin{equation}
\label{eq:QKV_linear_transformation}
Q^i = X^i_a \cdot W^i_Q, 
\quad
K^i = X^i_a \cdot W^i_K, 
\quad
V^i = X^i_a \cdot W^i_V 
\end{equation}

\begin{equation}
\label{eq:attention_output}
X_{O}^{i} = f_{\text{norm}}\left({f_{\text{Softmax}}\left( \frac{Q^{i} {K^{i}}^T}{\sqrt{h}} \right) \cdot V^{i} \cdot w_{O}^{i} + X^{i}}\right)
\end{equation}
For simplicity, we ignore memory fragmentation; analyzing the computational data flow of the attention layer (Eq. xxx), the total activation memory with half precision can be expressed as:
\begin{equation}
    \mathcal{M}_{\text{attn}} \approx (2B \cdot S \cdot n_h \cdot d_{\text{head}} + 
    2B \cdot S \cdot n_{kv} \cdot d_{\text{head}}) 
    \cdot 2 \text{ bytes}
    \tag{4}
\end{equation}
The first term accounts for storing the input hidden\_states $ X^i$ and $\mathbf{Q}$ tensors (Assume that the model's hidden dimension is equal to $n_h \times d_{head}$), 
while the second term accounts for the $K$ and $V$ tensors (i.e., KV caches).
Assuming $B = 1$, $S \approx 921600$, $d_{\text{model}} = 3584$, $d_{\text{head}} = 128$, $n_h = 28$, $n_{kv} = 4$, we compute:
\begin{align}
    \mathcal{M}_{\text{attn}} &= (2  \cdot 921600 \cdot 28 \cdot 128 
    +
    2  \cdot 921600 \cdot 4 \cdot 128)  \cdot 2 \\
    &= 15,099,494,400 \text{ bytes} \\
    &\approx \boxed{14.1 \text{ GB}}
\end{align} 
With chunked-prefill using $C = 4096$ chunk size, we can reduce the sequence length $S$ by $C = 225$ times, reducing $\mathcal{M}_{\text{attn}}$ from 17.6 GB to approximately 0.08 GB. This dramatic reduction enables the processing of extremely long sequences that would otherwise be infeasible.

\noindent\textbf{\emph{MLP Layer.} }
The SwiGLU (Swish-Gated Linear Unit) enhances transformer models through improved gating mechanisms and has been adopted as the default MLP architecture in many popular LLMs including InternVL2.5 and Qwen2.5 series. For input tensor $X^i$, the SwiGLU operation is defined as:

\begin{equation}
\label{eq:next_step_output}
X^{i+1} = f_{\text{silu}}\left( X_{O}^{i} \cdot W_{1} \right) * \left(W_2 \cdot X^i_O \right) \cdot W_{3} + X_{O}^{i}
\end{equation}

For a batch of sequences, activation memory analysis reveals requirements at each computational step. With batch size $B$, sequence length $S$, hidden dimension $d_{\text{model}}$, intermediate dimension $d_{\text{ff}}$, and data type float16, the total activation memory for a single SwiGLU layer is:

\begin{equation}
    \mathcal{M}_{\text{act}} = (B \cdot S \cdot (2d_{\text{model}} + 3d_{\text{ff}})) \cdot 2 \text{ bytes}
    \tag{9}
\end{equation}

For a one-hour video sampled with 1 FPS (3600 frames in total), parameters can be set $B = 1$, $S  \approx 921600$, $d_{\text{model}} = 4096$, and $d_{\text{ff}} = 14336$:

\begin{align}
    \mathcal{M}_{\text{act}} &= (1 \cdot 921600 \cdot (2 \cdot 4096 + 3 \cdot 14336)) \cdot 2  \\
    &= 94,371,840,000 \text{ bytes} \\
    &\approx \boxed{87.9 \text{ GB}}
\end{align}

This substantial memory requirement highlights the computational challenges in deploying SwiGLU-based models for high-resolution inputs with extended sequence lengths. However, if we prefill the tokens chunk by chunk, we can reduce the $S$ by $C$ times, and thus reduce the activation memory $\mathcal{M}_{\text{act}}$ by $C$ times. Assuming each chunk contains 4096 visual tokens , then $C = \frac{921600}{4096} = 225$ and we can reduce $\mathcal{M}_{\text{act}}$ from 87.9 GB to 0.4 GB.

\noindent\textbf{KV Cache Memory Analysis. }
When using Qwen2.5-VL-7B, with $|V| = 921600$ visual tokens, $|Q| = 256$ text tokens, $L = 28$ layers, $n_{kv} = 4$ heads, and $d_h = 128$, the total memory required to store the KV cache in half precision is:

\begin{equation}
\begin{split}
    \text{Memory} &= 2 \times L \times (|V| + |Q|) \times n_{\text{kv}} \times d_h \times 2 \\
                  &=52,862,910,464 \text{ bytes} \\
                  &\approx \boxed{49.2 \text{ GB}}.
\end{split}
\tag{13}
\end{equation}

 
 

\begin{table}[t]
    \small
    \centering
    \begin{adjustbox}{width=0.95\columnwidth}
    \begin{tabular}{lr}
    \toprule
    \textbf{Hyper-parameter} & \textbf{Value} \\
    \midrule
    \multicolumn{2}{l}{{\textit{Visual Encoder}}} \\
    \midrule
    Frame Sampling Rate &FPS=1 \\
    Input Resolution & 224*224 \\
    Visual Tokens per Image & 32 \\
    Patch Size & 14x14 \\
    \midrule

    \multicolumn{2}{l}{{\textit{Large Language Model}}} \\
    \midrule
    Number of Layers & 28 \\
    Hidden Size & 3584  \\
    Vocabulary Size & 152064 \\
    Number of Attention Heads & 28 \\
    Number of KV Heads & 4 \\
    \midrule
    
    \multicolumn{2}{l}{{\textit{Streaming KV-cache}}} \\
    \midrule
    $\alpha$ & 3 \\
    Maximum number of retrievals & 256 \\
    GPU-Memory(A100 SXM4) & 80GB \\

    \bottomrule
    \end{tabular}
    \end{adjustbox}
    \caption{The implementation settings of the evaluation.}
    \label{tab:supp_hyperpara}
\end{table}

\begin{table}[t]
    \small
    \centering
    \begin{adjustbox}{width=0.95\columnwidth}
    \begin{tabular}{lr}
    \toprule
    \textbf{Hyper-parameter} & \textbf{Value} \\
    \midrule
    \multicolumn{2}{l}{{\textit{Visual Encoder}}} \\
    \midrule
    Frame Sampling Rate &FPS=0.5 \\
    Input Resolution & 448*448 \\
    Visual Tokens per Image & 128 \\
    Patch Size & 14x14 \\
    \midrule

    \multicolumn{2}{l}{{\textit{Large Language Model}}} \\
    \midrule
    Number of Layers & 28 \\
    Hidden Size & 3584  \\
    Vocabulary Size & 152064 \\
    Number of Attention Heads & 28 \\
    Number of KV Heads & 4 \\
    \midrule
    
    \multicolumn{2}{l}{{\textit{Streaming KV-cache}}} \\
    \midrule
    $\alpha$ & 3 \\
    Maximum number of retrievals & 256 \\
    GPU-Memory(A100 SXM4) & 80GB \\
    \midrule
    
    \multicolumn{2}{l}{{\textit{ReKV}}} \\
    \midrule
    Top-k & 64 \\
    Block Size & 1 \\
    GPU-Memory(A100 SXM4) & 80GB \\

    \bottomrule
    \end{tabular}
    \end{adjustbox}
    \caption{The implementation settings of the streaming KV-cache ablation study.}
    \label{tab:supp_main_hyperpara}
\end{table}

\section{Detail of Evaluation}
In this section, we analyze the memory consumption of our approach, focusing on activation memory and key-value (KV) cache memory.

\noindent\textbf{Implement Setting}

We present the detail of the hyperparameters during the inference in Table~\ref{tab:supp_main_hyperpara}.

\noindent\textbf{Evaluation Result}

We present the detail of the evaluation results on StreamingBench in Table~\ref{tab:streamingbench}. StreamAgent achieves state-of-the-art performance among open-source models with an overall score of 57.02, outperforming the recent
online model Dispider-7B by 3.90 points (57.02 vs. 53.12).
While proprietary model Gemini 1.5 pro leads with 67.07. 

\noindent\textbf{Ablation Setting}

In the experiment of the ablation in streaming KV-cache, we use FPS=0.5 following the setting in ReKV. Notably, we evaluate the streaming KV-cache in offline setting aglining with ReKV, the detail of the hyperparameters are presented in Table~\ref{tab:supp_hyperpara}.
\section{Pseudocode}

\subsection{Algorithm of the Streaming KV-cache.}

We present the algorithm of the streaming KV-cache in Algorithm~\ref{alg:kv} 

\begin{algorithm}[!t]
\caption{Streaming KV-cache Algorithm}
\label{alg:kv}
\begin{algorithmic}[1]
\Require Incoming video stream $\{v_1, v_2, \dots, v_T\}$, query $Q_t$, threshold margin $\alpha$
\Ensure Output answer with retrieved memory-aware KV-cache
\State Initialize long-term memory in CPU $\mathcal{M}_\text{long} \gets \emptyset$
\For{each video clip $v_i$}
    \State Divide $v_i$ into chunks $\mathbf{C}^{v_i} = \{\mathbf{Z}_j^{v_i}\}_{j=1}^{\mu}$
    \State Initialize short-term KV-cache $H_\text{chunk}^{k,v} \gets \emptyset$
    \For{each chunk $X$ in $\mathbf{C}^{v_i}$}
        \State Compute current $K, V$ using Eq.~\eqref{eq:QKV_linear_transformation}
        \State 
        Prefill current chunk X
        \State
        Append $(K,V)$ to $H_\text{chunk}^{k,v}$
    \EndFor
    \State Store $H_\text{chunk}^{k,v}$ into long-term memory: $\mathcal{M}_\text{long} \gets \mathcal{M}_\text{long} \cup H_\text{chunk}^{k,v}$
\EndFor
\Statex
\State \textbf{Selective Recall}
\State Compute average query attention descriptor: $\mathbf{q}_{\text{avg}} = \frac{1}{T_q} \sum_{j=1}^{T_q} \mathbf{q}_j$
\State Initialize retrieved KV-cache $H_\text{out}^{k,v} \gets \emptyset$
\For{each layer $l$}
    \State Compute frame-wise attention scores $\mathbf{S}_h$ between $\mathbf{q}_{\text{avg}}$ and stored $\mathbf{k}_j$
    \State Select top frames: $\mathcal{J}_h = \{ j \mid \max(\mathbf{S}_h) - \mathbf{S}_{h,j} \leq \alpha \}$
    \State Append corresponding $k_j, v_j$ to $H_\text{out}^{k,v}$
\EndFor
\State Compute final $K,V$ with $H_\text{out}^{k,v}$ and $X_q$
\State \Return Answer generated via attention over $(K,V)$
\end{algorithmic}
\end{algorithm}


\section{Future Work}

In the future, we aim to use this agent-based workflow to construct data, and then fine-tune models using the constructed data to enhance their tool-using capabilities and their ability to plan for future events. We also intend to apply this approach in real-world scenarios. This planning-then-tool-invocation framework can naturally transfer to practical applications, for example, in intelligent surveillance systems, cameras can proactively adjust their monitoring areas or focal lengths based on the agent's planning to focus on specific regions. Additionally, we hope to leverage more tools specifically designed for video understanding. For instance, when counting repetitions, MLLMs often tend to overcount or undercount. We can use optical flow or other techniques to assist large models in achieving more accurate counting.

\end{document}